\documentclass[conference]{IEEEtran}
\usepackage{xcolor}
\usepackage{amsmath}
\usepackage{graphicx}
\usepackage{cite}
 \usepackage{amsfonts}
\usepackage[ruled,vlined]{algorithm2e}
\usepackage{url}
 \usepackage{multirow}
\usepackage{subcaption} 

\usepackage[table]{xcolor}
\usepackage[graphicx]{realboxes}
\usepackage{titlesec}
\usepackage{booktabs}
\makeatletter
\makeatletter
\def\ps@IEEEtitlepagestyle{%
  \def\@oddfoot{\mycopyrightnotice}%
  \def\@evenfoot{}%
}
\def\mycopyrightnotice{%
  {
  }
  \gdef\mycopyrightnotice{}
}

\makeatother
\title{
RPG-AE: Neuro-Symbolic Graph Autoencoders with Rare Pattern Mining for Provenance-Based Anomaly Detection.
}
\author{
   \IEEEauthorblockN{Asif Tauhid, Sidahmed Benabderrahmane, Mohamad Altrabulsi, Ahamed Foisal, Talal Rahwan}
    \IEEEauthorblockA{
    New York University, NYUAD, Computer Science Department,
    Division of Science.\\
    sidahmed.benabderrahmane@nyu.edu}
}
\usepackage[justification=raggedright,singlelinecheck=false]{caption}

\begin{document}

\maketitle

%
\begin{abstract} 
Advanced Persistent Threats (APTs) are sophisticated, long-term cyberattacks that are difficult to detect because they operate stealthily and often blend into normal system behavior. This paper presents a neuro-symbolic anomaly detection framework that combines a Graph Autoencoder (GAE) with rare pattern mining to identify APT-like activities in system-level provenance data. Our approach first constructs a process behavioral graph using k-Nearest Neighbors based on feature similarity, then learns normal relational structure using a Graph Autoencoder. Anomaly candidates are identified through deviations between observed and reconstructed graph structure. To further improve detection, we integrate an rare pattern mining module that discovers infrequent behavioral co-occurrences and uses them to boost anomaly scores for processes exhibiting rare signatures. We evaluate the proposed method on the DARPA Transparent Computing datasets and show that rare-pattern boosting yields substantial gains in anomaly ranking quality over the baseline GAE. Compared with existing unsupervised approaches on the same benchmark, our single unified model consistently outperforms individual context-based detectors and achieves performance competitive with ensemble aggregation methods that require multiple separate detectors. These results highlight the value of coupling graph-based representation learning with classical pattern mining to improve both effectiveness and interpretability in provenance-based security anomaly detection.\\
\textbf{Code}: https://github.com/asiftauhid/gae\_anomaly\_detection
\end{abstract}
\section{Introduction}
Modern cybersecurity threats have become increasingly sophisticated, rendering traditional signature-based detection insufficient for defending complex systems \cite{dehankar2026machine,al2025deep}. Advanced Persistent Threats (APTs) are particularly challenging due to their stealthy, long-term nature and their ability to evade conventional defenses through behaviors that closely resemble legitimate activity \cite{al2026dynamic,benabderrahmane2025apt,benabderrahmane2025ranking}. Detecting such attacks therefore requires approaches that can model not only individual events, but also the relationships and dependencies among system entities over time \cite{benabderrahmane2025flagus,benabderrahmane2025one}.

System provenance data, which records causal interactions among processes, files, and network connections, provides a rich substrate for security analytics because it preserves behavioral context and dependency structure \cite{wang2025dynamic}. However, provenance traces are typically high-dimensional and structurally complex, making them difficult to analyze using classical machine learning techniques that assume independent samples or shallow feature interactions \cite{wu2025angus}. Graph Neural Networks (GNNs), and in particular Graph Autoencoders (GAEs), offer a natural framework for learning compact representations from relational data and for detecting anomalies via deviations from learned normal structure \cite{luo2025conformal}.

Despite this promise, provenance-based anomaly detection remains difficult in practice due to several intertwined challenges: (1) high-dimensional behavioral spaces where activity is described by many sparse indicators; (2) structural complexity arising from heterogeneous and densely connected interactions; (3) the inherently rare nature of attack behaviors, which may appear as infrequent co-occurrence signatures that are easily diluted by dominant benign patterns; and (4) extreme class imbalance, where normal processes vastly outnumber attack instances \cite{wu2025angus}. These factors often lead purely representation-learning approaches to under-emphasize weak but informative rare signals, while purely pattern-based methods can struggle to capture relational context.

To address these challenges, we propose \emph{RPG-AE} a hybrid neuro-symbolic framework that combines Graph Autoencoders with rare pattern mining for provenance-based anomaly detection. Our approach constructs a process behavioral graph using k-Nearest Neighbors based on feature similarity, learns normal relational structure via a GAE, and identifies anomalies using reconstruction-based deviations. To strengthen detection of subtle, infrequent attack signatures, we additionally mine rare behavioral co-occurrence patterns using an Apriori-based procedure and incorporate this information through a score boosting mechanism that augments the GAE anomaly scores with rarity evidence.

Our main contributions are summarized as follows:
\begin{itemize}
    \item Graph autoencoder for provenance behavior: We design a GAE-based anomaly detector that learns normal behavioral structure from system provenance data using k-NN graph construction and GCN-based encoding.
    \item Rare pattern augmentation: We integrate an Apriori-based rare pattern mining module to capture infrequent behavioral combinations that may indicate malicious activity.
    \item Boosted scoring mechanism: We introduce a fusion strategy that combines GAE reconstruction error with rare-pattern evidence to improve anomaly ranking.
    \item Empirical evaluation: We perform a comprehensive evaluation on the DARPA Transparent Computing datasets, comparing against existing unsupervised baselines.
\end{itemize} 

\section{Related work}
\label{sec:related}

\paragraph{APT Detection and System Provenance:} Advanced Persistent Threats (APTs) pose unique challenges for detection due to their stealthy, long-term nature and their tendency to blend with legitimate system activity. Alshamrani et al.~\cite{alshamrani2019apt} survey APT techniques and emphasize the limitations of signature-based defenses, motivating anomaly detection methods that can capture subtle and persistent deviations. System provenance data records causal dependencies among system entities (e.g., processes, files, and network connections), providing rich behavioral context for detecting sophisticated attacks~\cite{berrada2020baseline}.

The DARPA Transparent Computing (TC) program has served as a major benchmark for provenance-based security analytics. Berrada et al.~\cite{berrada2020baseline} established a widely used unsupervised evaluation baseline on DARPA TC by studying multiple classical detectors across operating systems and analysis contexts. Other provenance-driven systems include UNICORN~\cite{han2020unicorn}, which uses provenance sketching for runtime detection, StreamSpot~\cite{manzoor2016streamspot} for memory-efficient anomaly detection in streaming heterogeneous graphs, and SLEUTH~\cite{hossain2017sleuth} for real-time attack scenario reconstruction from audit data. These systems demonstrate the value of provenance graphs for security monitoring and investigation, but they are often oriented toward streaming constraints, scenario reconstruction, or context-specific detection pipelines. In contrast, our focus is a single unified unsupervised anomaly ranking model that combines graph representation learning with complementary symbolic evidence.

\paragraph{Graph Neural Networks for Anomaly Detection}
Graph Neural Networks (GNNs) have shown strong performance in learning representations from relational data across domains such as social networks, molecules, and recommender systems. Ma et al.~\cite{ma2023survey} provide a comprehensive survey of deep graph anomaly detection, highlighting how architectures such as Graph Convolutional Networks (GCNs), Graph Attention Networks (GATs), and Graph Autoencoders (GAEs) can capture contextual and structural irregularities. Mir et al.~\cite{mir2023gcn} further demonstrate that GCN-based models can learn discriminative node representations for detecting anomalous behavior in graph-structured settings.

Graph Autoencoders extend classical autoencoders to graphs by learning low-dimensional embeddings that preserve graph structure, and then reconstructing connectivity (or edge probabilities) from these embeddings. The variational graph autoencoder framework introduced by Kipf and Welling~\cite{kipf2016vgae} has become a foundational approach for unsupervised graph representation learning and has been adapted to a range of anomaly detection scenarios. While reconstruction-based objectives are effective for modeling dominant (frequent) structure, they may under-emphasize weak but informative signals under extreme class imbalance---a common characteristic of system provenance traces. Our work follows the GAE paradigm for modeling normal relational structure, and complements it with rare co-occurrence evidence to increase sensitivity to infrequent yet security-relevant behaviors.

\paragraph{Association Rule and Pattern Mining for Security:}
Association rule mining and frequent itemset discovery are classical data mining techniques for extracting co-occurrence structure from large transactional datasets. The Apriori algorithm~\cite{agrawal1994apriori} remains a foundational approach and has been widely used in security settings for tasks such as alert correlation, attack pattern discovery, and log-driven behavioral analysis. In the provenance domain, Benabderrahmane et al.~\cite{benabderrahmane2021vrarm} proposed VR-ARM, a rule-mining-based detection method that leverages both frequent and rare association rules to score process anomalousness in system traces.

Pattern- and rule-based detectors offer interpretability and can be effective at highlighting infrequent behavioral combinations. However, they typically operate in transaction space and do not explicitly learn a continuous latent manifold that captures relational similarity among processes. Our approach bridges these perspectives by coupling a GAE-based structural model with Apriori-mined rare co-occurrence patterns. Rather than using mined patterns as a standalone detector, we incorporate them as auxiliary evidence to strengthen reconstruction-based anomaly scoring, yielding a hybrid graph--pattern framework that is both effective and interpretable.

\section{Proposed Framework}
\label{sec:framework}
\begin{figure}
    \centering
    \includegraphics[width=0.8\linewidth]{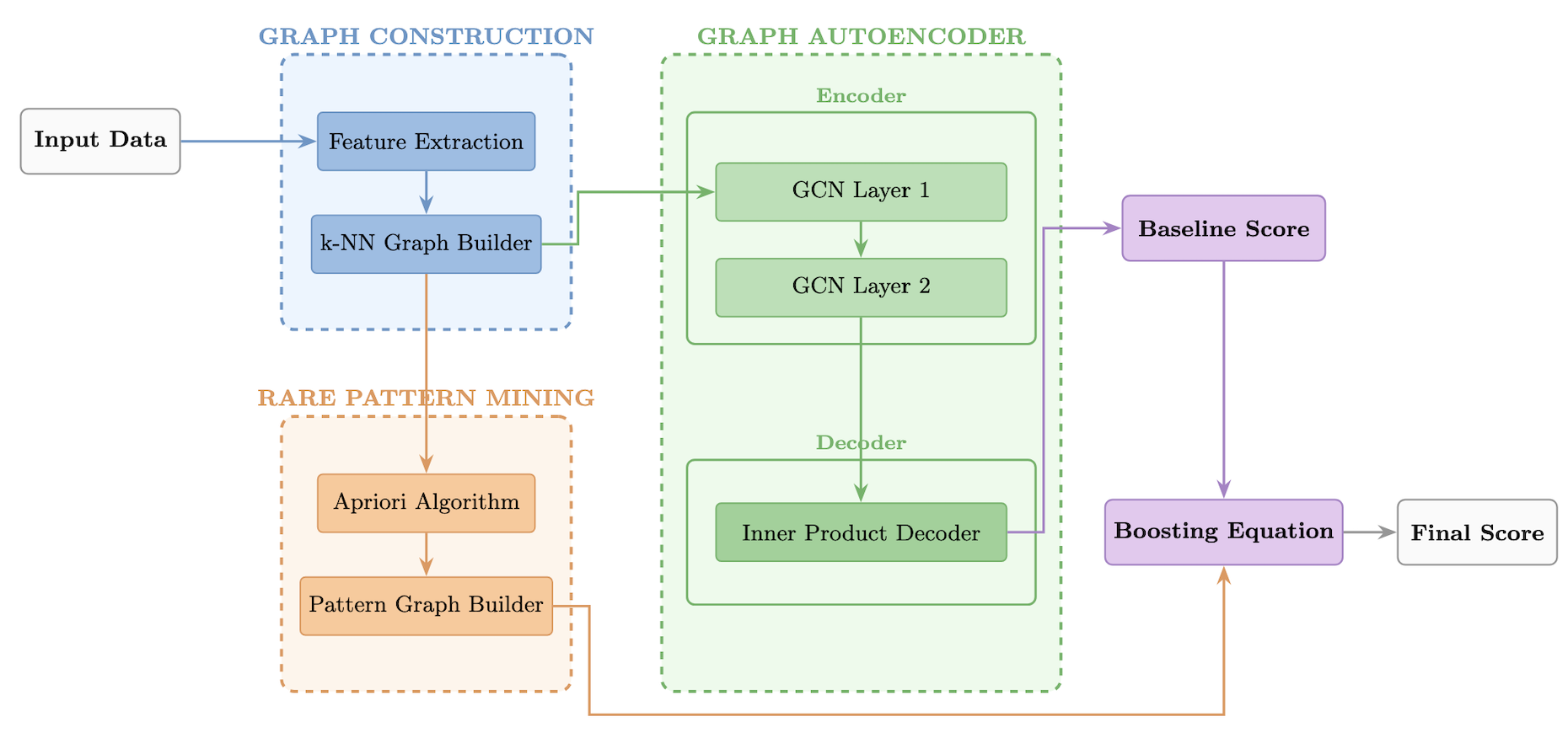}
    \caption{Overview of the proposed RPG-AE architecture. The pipeline first extracts process-level behavioral features from provenance logs and constructs a k-NN process similarity graph. In parallel, an Apriori-based rare pattern mining module discovers infrequent co-occurrence patterns and converts them into pattern evidence. A Graph Autoencoder is trained to learn normal relational structure. At inference, processes are scored using graph reconstruction residuals and then boosted with rare-pattern evidence to produce the final anomaly ranking.}
    \label{fig:arch}
\end{figure}

\begin{figure}
    \centering
    \includegraphics[width=0.7\linewidth]{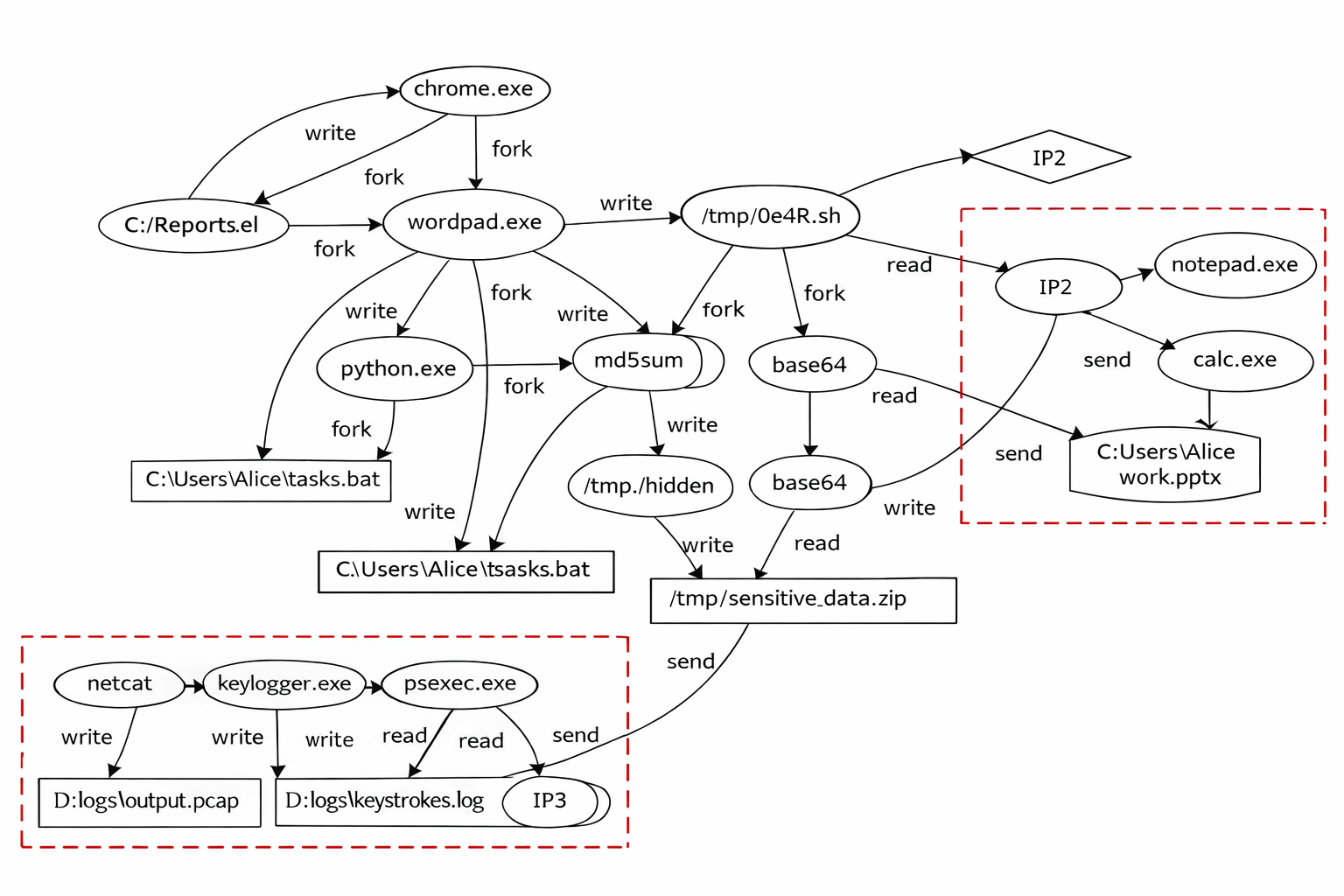}
    \caption{Example system provenance graph. Nodes represent system entities—processes (ovals), files (rectangles), and network endpoints (diamonds)—while directed edges encode causal interactions such as read, write, fork/execute, send, and receive. The dashed red boxes highlight subgraphs exhibiting suspicious behavior patterns (e.g., uncommon process spawning chains, sensitive file access, and outbound communication), illustrating the relational structure exploited by our rare-pattern-guided graph autoencoder for anomaly detection.}
    \label{fig:provenance}
    \vspace{-1 em}
\end{figure}

%
\paragraph{Overview:}
\label{subsec:overview}
Figure~\ref{fig:arch} summarizes the proposed RPG-AE framework, which consists
of five main components: input data processing, graph construction with rare pattern mining, graph Autoencoder, score computation combining baseline
and boosting equation, and final anomaly score output. Given system-level provenance logs, we derive a process-centric representation in which each process is described by a behavioral feature vector and an equivalent transaction view. Using these features, we construct a process similarity graph via $k$-Nearest Neighbors (k-NN) and learn a model of \emph{normal} relational structure with a Graph Autoencoder (GAE) trained for error reconstruction. In parallel, we mine \emph{rare} behavioral co-occurrence patterns using an Apriori-style procedure and use the resulting symbolic evidence to refine (boost) reconstruction-based anomaly scores. The final output is an anomaly ranking over processes, together with rare patterns that can serve as interpretable indicators for flagged behaviors.
\paragraph{Provenance Data:}
Figure~\ref{fig:provenance} illustrates the type of relational evidence captured by system provenance and motivates our modeling choices. Provenance naturally forms a directed, heterogeneous interaction graph in which a single process may simultaneously interact with files and remote endpoints, and where attack behaviors often appear as \emph{multi-hop} dependencies rather than isolated events. The highlighted subgraphs exemplify typical suspicious motifs, such as atypical process creation chains, unexpected access to sensitive artifacts, and outbound network communication following local collection activity. Importantly, these behaviors are rarely detectable from individual events alone: they emerge from the \emph{combination} of entity types and causal edges and from the \emph{co-occurrence} of otherwise benign actions in an unusual context. This observation underpins our framework: the graph autoencoder captures deviations in relational structure through reconstruction errors, while rare pattern mining emphasizes infrequent co-occurrence signatures that may be diluted in purely similarity-driven graph learning.

\paragraph{Notation:}
Let $\mathcal{P}=\{p_1,\dots,p_n\}$ denote the set of $n$ processes extracted from provenance data. Each process $p_i$ is represented by a $d$-dimensional behavioral vector $\mathbf{x}_i\in\{0,1\}^d$, and we stack these vectors into a feature matrix $X\in\mathbb{R}^{n\times d}$. We also view each process as a transaction $T_i=\{f_j \mid x_{ij}=1\}$ over a vocabulary of behavioral items/features. 

We construct a process graph $G=(V,E)$ where $V=\mathcal{P}$ and edges encode behavioral similarity. Specifically, for each node $p_i$, we compute its $k$ nearest neighbors $\mathcal{N}_k(i)$ under a similarity function $\mathrm{sim}(\cdot,\cdot)$ (e.g., cosine similarity), and define an adjacency matrix $A\in\{0,1\}^{n\times n}$ such that $A_{ij}=1$ if $p_j\in\mathcal{N}_k(i)$. We denote by $\tilde{A}$ the normalized adjacency used by the GCN encoder (e.g., with self-loops and symmetric normalization). The GAE encoder maps node features to latent embeddings $Z\in\mathbb{R}^{n\times h}$, where $\mathbf{z}_i$ is the embedding of process $p_i$, and the decoder produces a reconstructed adjacency $\hat{A}$ used for link prediction. We compute a baseline anomaly score $s_{\mathrm{gae}}(p_i)$ from reconstruction discrepancies, and a rare-pattern score $s_{\mathrm{rare}}(p_i)$ derived from the set of mined rare patterns supported by $T_i$. The final anomaly score $s(p_i)$ is obtained by combining these two signals, and processes are ranked in decreasing order of $s(\cdot)$.
\paragraph{Graph Construction:}
We construct a process similarity graph using a k-nearest neighbors (k-NN) strategy in the standardized feature space. For each process, its feature vector is used to compute the pairwise proximity using cosine distance. Each node is then connected to its k = 5 nearest neighbors, and edges are added bidirectionally to obtain an undirected behavioral graph. This construction captures local behavioral similarity by linking processes that exhibit comparable activity patterns.

\paragraph{Graph Autoencoder Architecture:}
\label{subsec:gae}

We employ a Graph Autoencoder (GAE) composed of a two-layer Graph Convolutional Network (GCN) encoder and an inner-product decoder. Given the node feature matrix $X \in \mathbb{R}^{n \times d}$ and the symmetrically normalized adjacency matrix with self-loops $\hat{A}$, the encoder computes
$H^{(1)} = \mathrm{ReLU}\!\left(\hat{A} X W^{(0)}\right)$ and $Z = \hat{A} H^{(1)} W^{(1)}$,
where $W^{(0)} \in \mathbb{R}^{d \times h_1}$ and $W^{(1)} \in \mathbb{R}^{h_1 \times h_2}$ are learnable parameters. The resulting embedding matrix $Z \in \mathbb{R}^{n \times h_2}$ provides a compact representation of process behavior that is used by the decoder to reconstruct graph connectivity.

The decoder reconstructs the adjacency matrix via inner products between node embeddings: $\hat{A}_{ij} = \sigma\!\left(z_i^\top z_j\right)$, where $\sigma(\cdot)$ is the sigmoid function and $z_i$ denotes the embedding of node $i$. The model is trained for unsupervised link prediction using a binary cross-entropy objective and optimized with Adam (learning rate $0.05$) for 200 epochs. At inference time, we score each node by its local reconstruction error, computed as the average discrepancy between observed and reconstructed edges in its neighborhood, and use this score for anomaly ranking.

\paragraph{Rare Pattern Mining:}
\label{subsec:rare_patterns}

To enhance sensitivity to infrequent but potentially malicious behaviors, we incorporate rare pattern mining using the Apriori algorithm. The procedure consists of three steps:

\noindent(1) Transaction conversion: We convert the binary feature matrix into a transactional representation suitable for itemset mining. Each process $i$ is treated as a transaction $T_i$ containing the indices (or tokens) of features that are active for that process, i.e., $T_i = \{f \mid X_{if}=1\}$.

\noindent(2) Rare pattern discovery: We apply Apriori to extract \emph{rare} co-occurrence patterns by constraining the support of candidate itemsets to lie within a predefined interval. Specifically, we retain patterns whose empirical support satisfies
$
min\_support = 0.012\% \;\le\; \mathrm{supp}(p) \;\le\; 1.22\% = max\_support,
$
thereby filtering out both extremely common patterns (likely benign background behavior) and vanishingly rare patterns (often noisy or non-informative).

\noindent(3) Pattern graph construction: Using the mined patterns, we construct a pattern-induced graph $G^{pat}=(V,E^{pat})$, in which two processes are connected if they share at least one rare pattern. This graph provides an additional relational signal that highlights uncommon behavioral similarities and is later used as auxiliary evidence in our anomaly scoring mechanism.
 \paragraph{Score Boosting Mechanism:}
\label{subsec:boosting}

We enhance the reconstruction-based anomaly scores produced by the GAE using a degree-based boosting strategy derived from the rare-pattern graph. Let $s_{\text{base}}(i)$ denote the baseline anomaly score for node (process) $i$ obtained from GAE reconstruction. We define the boosted score as
\begin{equation}
s_{\text{boosted}}(i) = s_{\text{base}}(i) + \alpha \cdot b(i),
\label{eq:boosting}
\end{equation}
where $b(i)$ is a normalized degree bonus (boosting factor) computed from the rare-pattern graph. Specifically,
\begin{equation}
b(i) = \frac{d(i)}{d_{\max}},
\label{eq:degree_bonus}
\end{equation}
with $d(i)$ denoting the degree of node $i$ in the rare-pattern graph and $d_{\max} = \max_j d(j)$ the maximum degree. Intuitively, processes that share rare behavioral patterns with many other processes receive a larger bonus, reflecting stronger evidence of unusual co-occurrence structure.
We tune the boosting weight $\alpha$ over the set $\{0.1, 0.5, 1.0, 2.0, 5.0, 10.0\}$ and select $\alpha = 2.0$ based on validation performance.

%
\paragraph{Workflow Summary:}
\label{subsec:workflow_summary}
Algorithm~\ref{alg:workflow} summarizes the end-to-end workflow of our RPG-AE framework. Starting from system provenance logs, we extract process-level behavioral features and construct a process similarity graph using a $k$-nearest neighbors strategy. In parallel, we mine rare behavioral co-occurrence patterns with Apriori and use the resulting patterns to build a rare-pattern graph that connects processes sharing at least one rare pattern. We then train a graph autoencoder on the similarity graph to learn normal relational structure via unsupervised link reconstruction and compute a reconstruction-based anomaly score for each process. Finally, we boost the baseline scores with degree-based evidence from the rare-pattern graph to obtain a ranked list of anomalous processes.

\begin{algorithm}[t]
\small
\caption{Workflow of the proposed RPG-AE framework}
\label{alg:workflow}
\KwIn{Provenance logs; graph parameter $k$; Apriori thresholds $(\texttt{min\_support}, \texttt{max\_support})$; boosting weight $\alpha$.}
\KwOut{Anomaly scores $s_{\text{boosted}}(i)$ and ranked anomalous processes.}

\BlankLine
\textbf{Feature extraction:} Convert provenance logs into process-level feature matrix $X \in \mathbb{R}^{n \times d}$.\\
\textbf{Similarity graph:} Standardize $X$ and build $k$-NN process graph $G^{knn}=(V,E^{knn})$ using cosine distance.\\
\textbf{Rare pattern mining:} Convert each process row into a transaction $T_i=\{f \mid X_{if}=1\}$ and mine rare patterns $\mathcal{P}$ with Apriori under $(\texttt{min\_support}, \texttt{max\_support})$.\\
\textbf{Rare-pattern graph:} Build $G^{pat}=(V,E^{pat})$ by connecting processes that share at least one pattern in $\mathcal{P}$.\\
\textbf{GAE training:} Train a GAE on $G^{knn}$ to reconstruct edges and obtain embeddings $Z$ and reconstructed adjacency $\hat{A}$.\\
\textbf{Baseline scoring:} Compute reconstruction-based anomaly score $s_{\text{base}}(i)$ for each process.\\
\textbf{Boosting:} Compute degree bonus $b(i)=d(i)/d_{\max}$ in $G^{pat}$ and output $s_{\text{boosted}}(i)=s_{\text{base}}(i)+\alpha b(i)$.\\
\Return{$\{s_{\text{boosted}}(i)\}_{i=1}^n$ and anomaly ranking.}
\end{algorithm}

%
\section{Experimental settings, results and analysis}
\paragraph{Datasets:}
\label{subsec:datasets}

The datasets used in this study are derived from DARPA's \texttt{Transparent Computing (TC)} program\footnote{\url{https://www.darpa.mil/program/transparent-computing}} and are processed using the ADAPT\footnote{\url{https://gitlab.com/adaptdata}} (Automatic Detection of Advanced Persistent Threats) ingester. The resulting corpus spans multiple operating systems, namely Android, Linux, BSD, and Windows, and includes two representative attack scenarios~\cite{DBLP:journals/fgcs/BenabderrahmaneHVCR24}. The code and configuration files necessary to reproduce our experiments are publicly available in the project repository\footnote{https://github.com/asiftauhid/gae\_anomaly\_detection}.

The provenance records are first transformed into graph databases and then subjected to integration and deduplication. For learning and evaluation, each dataset is represented in a process-centric tabular form: each row corresponds to a process and is encoded as a high-dimensional Boolean feature vector, where a value of $1$ indicates the presence of a particular behavioral attribute or event associated with that process.  Similar to prior work on DARPA TC, the resulting data are highly imbalanced (BSD: 13 anomalies /75,903 benign; Linux: 8/17,569; Windows: 25/247,160; Android: 9/102), with attack-related processes forming a very small fraction of all observed processes.

\begin{figure}
    \centering
    \includegraphics[width=0.65\linewidth]{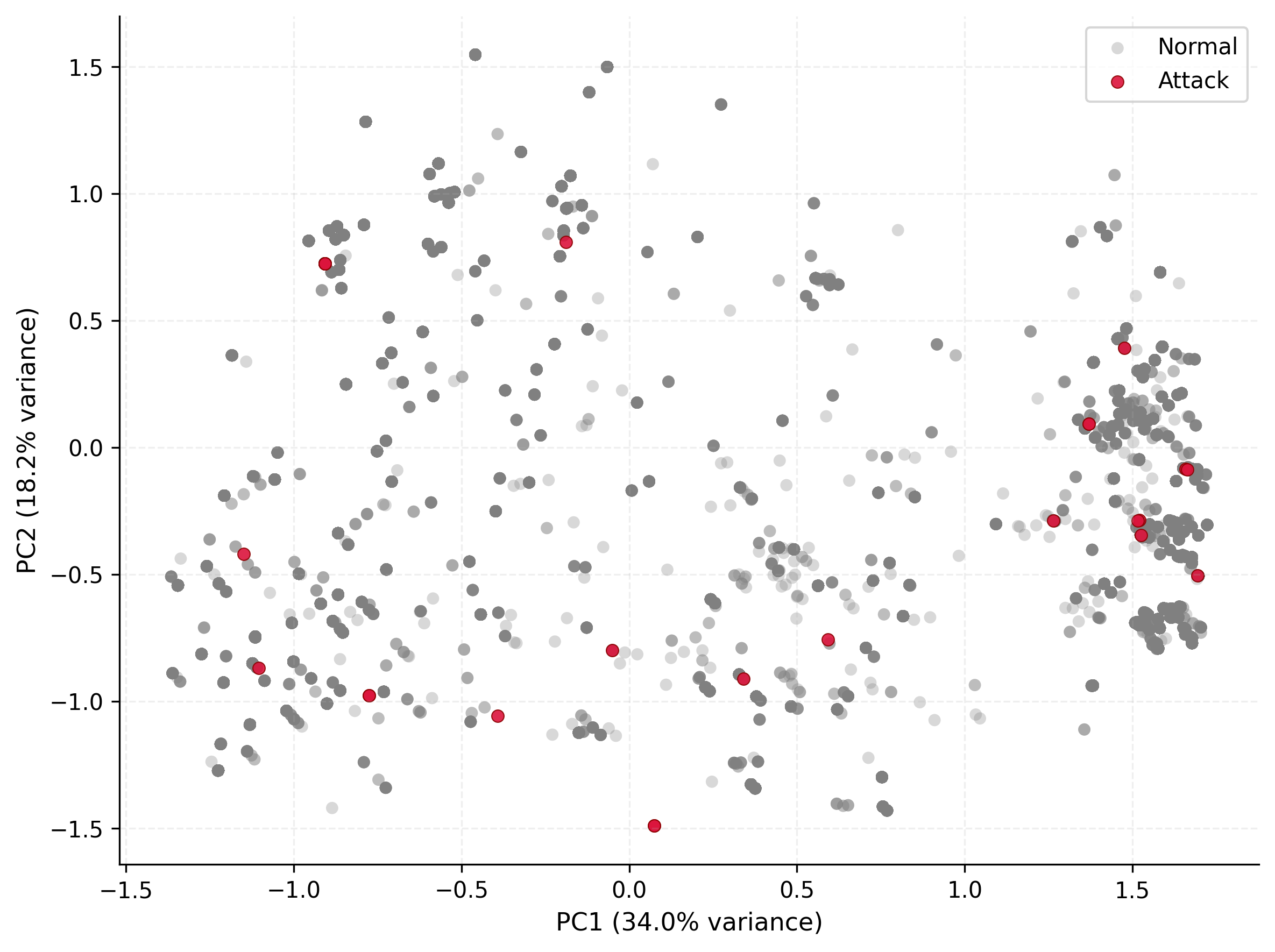}
    \caption{PCA visualization of process behavioral feature space for Linux Dataset. We project all processes, each described by 299 behavioral features, into two dimensions using Principal Component Analysis (PCA) for visualization. The first two components (PC1 and PC2) together capture approximately 52\% of the total variance, highlighting the overall distribution of processes in the reduced space (with attack instances shown in red and normal instances in gray).}
    \label{fig:scatter}
\end{figure}

Figure \ref{fig:scatter} visualizes the process-level behavioral space for Linux data, after projecting all processes (each represented by 299 features) onto the first two principal components. Although PC1 and PC2 together capture roughly 52\% of the variance, the plot shows substantial overlap between normal and attack processes, with attack points dispersed across regions dominated by benign behavior rather than forming a clearly separable cluster. This highlights a key challenge of provenance-based detection under extreme class imbalance: malicious activity often manifests as subtle deviations that remain close to normal operating patterns in a low-dimensional view. The absence of a clean boundary in the PCA projection motivates our use of relational modeling and rare-pattern evidence—rather than relying on feature-space separation alone—to prioritize suspicious processes based on structural inconsistency and infrequent behavioral co-occurrences.
\paragraph{Evaluation Metric:}
\label{subsec:metric}
Although the area under the ROC curve (AUC) is widely used for binary classification, it can be less informative in provenance-based detection due to \emph{extreme class imbalance}, where benign processes outnumber attack processes by several orders of magnitude. In such settings, a detector may achieve a strong AUC while still failing to prioritize the few truly malicious processes near the top of the output list, which is the operational requirement in security triage. We therefore evaluate performance from a \emph{ranking} perspective, aiming to surface the most suspicious processes first. To measure ranking quality, we use the normalized Discounted Cumulative Gain (nDCG), a standard metric in information retrieval \cite{benabderrahmane2021vrarm}. nDCG rewards methods that rank relevant (attack) processes early, while progressively discounting gains at lower ranks, reflecting the diminishing utility of late retrieval in practical investigations. The metric is normalized by the score of an ideal (perfect) ranking, yielding values between 0 and 1 and enabling fair comparison across datasets and attack scenarios.

\begin{figure}
    \centering
    \includegraphics[width=0.8\linewidth]{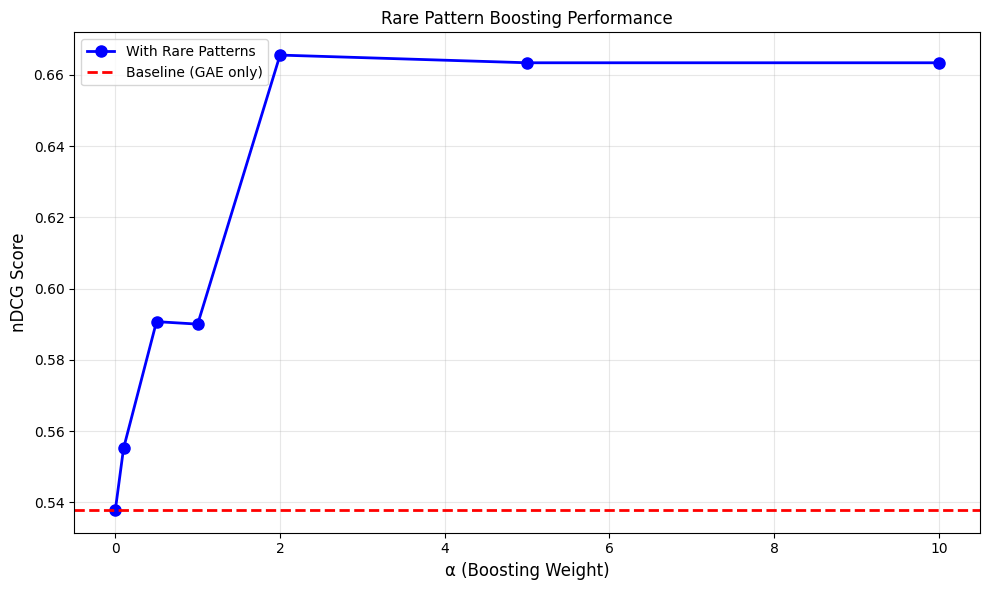}
    \caption{Effect of rare-pattern boosting on nDCG.}
    \label{fig:rareboosting}
    \vspace{-1 em}
\end{figure}

\paragraph{Performance Results:}
\label{subsec:perf_results}
\begin{figure}
    \centering
    \includegraphics[width=1\linewidth]{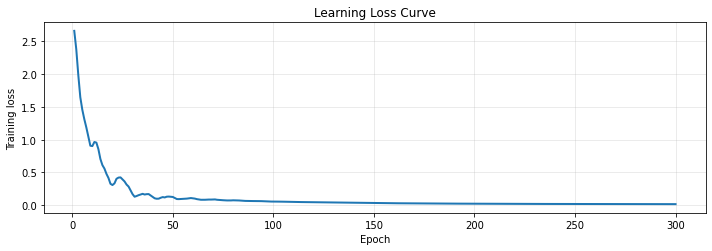}
    \caption{Training loss curve of RPG-AE (BSD dataset). The loss drops sharply during the first epochs, exhibits mild early oscillations, and then gradually stabilizes to a low plateau, indicating convergence.}
    \label{fig:learningloss}
\end{figure}
Figure \ref{fig:learningloss} presents the training loss curve of RPG-AE (BSD data). The early fluctuations reflect initial optimization instability, while the later flat region suggests the model has learned a stable reconstruction of the normal graph structure.
Table~\ref{tab:boosting} (together with Figure~\ref{fig:rareboosting}) reports the impact of incorporating rare-pattern evidence through the boosting weight $\alpha$.
When $\alpha=0$ (ablation), the model reduces to the baseline GAE that ranks processes purely using reconstruction-based deviations.
As $\alpha$ increases from small values (e.g., $\alpha=0.1$) to moderate values (e.g., $\alpha \in \{0.5,1.0,2.0\}$), the ranking performance improves steadily, indicating that the rare-pattern signal provides complementary information beyond structural reconstruction alone.
The best performance is obtained at $\alpha=2.0$, suggesting that a moderate emphasis on rare-pattern connectivity most effectively amplifies truly suspicious processes without overwhelming the baseline reconstruction signal.
For larger values (e.g., $\alpha \ge 5.0$), performance saturates and slightly declines, consistent with the intuition that overly strong boosting may start to favor highly connected rare-pattern neighborhoods even when their structural deviations are weak.
Overall, these results validate the benefit of integrating rare-pattern evidence: the hybrid score consistently improves upon the GAE-only detector and yields the most accurate prioritization when the two signals are well-weighted.

\paragraph{Comparison with Baseline Methods:}
\label{subsec:baseline_comparison}
Table~\ref{tab:baselines} summarizes the anomaly-ranking performance of our method against representative baselines on the same DARPA TC provenance benchmark. Our proposed \emph{RPG-AE} consistently delivers strong performance across operating systems and achieves the best nDCG on \emph{Linux}, \emph{Windows}, and \emph{Android}, while matching the top result on \emph{BSD}. Specifically, rare-pattern guidance yields clear gains over the graph-only baseline (GNN), improving nDCG from 0.56 to 0.66 on Linux, from 0.48 to 0.62 on Windows, from 0.25 to 0.52 on BSD, and from 0.41 to 0.85 on Android. These improvements confirm that rare co-occurrence evidence complements reconstruction-based deviations and is particularly effective in highly imbalanced settings where attacks may be weakly separable in feature space.

Compared to prior families of detectors, RPG-AE also outperforms classical statistical/compression approaches and standard outlier detectors (e.g., Isolation Forest, OC-SVM, LOF), which generally lag behind on Linux/Windows/BSD and exhibit larger cross-OS variability. In addition, RPG-AE improves upon association-rule baselines (VR-ARM and VF-ARM), which can perform competitively in some operating systems but fail to generalize consistently across all environments (e.g., near-zero performance on Android). Overall, these results indicate that combining a unified graph autoencoding backbone with rare-pattern mining yields a robust and portable detector that generalizes across diverse operating systems.

Beyond aggregate accuracy, Table~\ref{tab:baselines} highlights a practical advantage of our design: unlike approaches that rely on multiple context-specific detectors or ensemble aggregation, RPG-AE provides a single end-to-end framework where symbolic rarity evidence is injected directly into the anomaly scoring function. This integration simplifies deployment and supports interpretability, since elevated scores can be traced back to the rare behavioral patterns that contributed to the final ranking.



\begin{table}[t]
\centering
\caption{Effect of rare-pattern boosting on nDCG (with relative improvement).}
\label{tab:boosting}
\begin{tabular}{lccc}
\toprule
$\alpha$ & nDCG & Improvement & Notes \\
\midrule
0.0  & 0.5379 & --      & baseline GAE only \\
0.1  & 0.5552 & +3.2\%  & Modest improvement \\
0.5  & 0.5907 & +9.8\%  & Noticeable improvement \\
1.0  & 0.5900 & +9.7\%  & Similar to $\alpha=0.5$ \\
\textbf{2.0}  & \textbf{0.6656} & \textbf{+23.7\%} & \textbf{Best performance} \\
5.0  & 0.6634 & +23.3\% & Slight decline \\
10.0 & 0.6634 & +23.3\% & Plateaus \\
\bottomrule
\end{tabular}
\end{table}

\begin{table}[t]
\centering
\scriptsize
\caption{Comparison of nDCG scores on DARPA TC.}
\label{tab:baselines}

\begin{tabular}{|l|c|cccc|}
\hline
\multirow{2}{*}{\textbf{Method}} & \multirow{2}{*}{\textbf{Type}} & \multicolumn{4}{c|}{\textbf{nDCG Score}}                                                                                                \\ \cline{3-6} 
                                 &                                & \multicolumn{1}{c|}{\textbf{Linux}} & \multicolumn{1}{c|}{\textbf{Windows}} & \multicolumn{1}{c|}{\textbf{BSD}} & \textbf{Android}      \\ \hline
RPG-AE  (Ours)                         & GNN+Rare patterns              & \multicolumn{1}{c|}{\textbf{0.66}}           & \multicolumn{1}{c|}{\textbf{0.62}}                 & \multicolumn{1}{c|}{\textbf{0.52}}             &  \textbf{ 0.85 }                   \\ \hline 
GNN                              & Graph Baseline                       & \multicolumn{1}{c|}{0.56}           & \multicolumn{1}{c|}{0.48}                 & \multicolumn{1}{c|}{0.25}             &    0.41                   \\ \hline
VR-ARM \cite{benabderrahmane2021vrarm}                          & Rule mining                    & \multicolumn{1}{c|}{0.54}           & \multicolumn{1}{c|}{0.61}                 & \multicolumn{1}{c|}{0.36}             &   0                    \\ \hline
VF-ARM  \cite{benabderrahmane2021vrarm}                         & Rule mining                    & \multicolumn{1}{c|}{0.13}           & \multicolumn{1}{c|}{0.50}                 & \multicolumn{1}{c|}{0.18}             & 0                      \\ \hline
OC3 \cite{DBLP:journals/fgcs/BenabderrahmaneHVCR24}                             & Compression                    & \multicolumn{1}{c|}{0.41}           & \multicolumn{1}{c|}{0.43}                 & \multicolumn{1}{c|}{0.38}             &   0.82                    \\ \hline
AVF \cite{DBLP:journals/fgcs/BenabderrahmaneHVCR24}                             & Statistical                    & \multicolumn{1}{c|}{0.29}           & \multicolumn{1}{c|}{0.52}                 & \multicolumn{1}{c|}{\textbf{0.52}}             &    0.83                   \\ \hline
I-Forest\cite{DBLP:journals/fgcs/BenabderrahmaneHVCR24}                         & ML                             & \multicolumn{1}{c|}{0.17}           & \multicolumn{1}{c|}{0.12}                 & \multicolumn{1}{c|}{0.13}             & \multicolumn{1}{c|}{0.32} \\ \hline
OC-SVM\cite{DBLP:journals/fgcs/BenabderrahmaneHVCR24}                           & ML                             & \multicolumn{1}{c|}{0.17}           & \multicolumn{1}{c|}{0.14}                 & \multicolumn{1}{c|}{0.13}             & \multicolumn{1}{c|}{0.32} \\ \hline
LOF \cite{DBLP:journals/fgcs/BenabderrahmaneHVCR24}                             & ML                             & \multicolumn{1}{c|}{0.15}           & \multicolumn{1}{c|}{0.018}                 & \multicolumn{1}{c|}{0.07}             & \multicolumn{1}{c|}{0.18} \\ \hline
DeepOCCL \cite{wang2023deep} &Deep Learning    & \multicolumn{1}{c|}{0.20 }                 & \multicolumn{1}{c|}{0.26}             & \multicolumn{1}{c|}{0.22 } &0.75\\ \hline
\end{tabular}
\end{table}


\section{Conclusion}
\label{sec:conclusion}

This paper introduced a hybrid neuro-symbolic framework for provenance-based anomaly detection that combines graph representation learning with rare behavioral pattern mining. We construct a process similarity graph from provenance-derived features and learn normal relational structure using a GAE trained via unsupervised link reconstruction. To improve sensitivity to infrequent but security-relevant behaviors, we mine rare co-occurrence patterns with Apriori and inject this evidence through a degree-based score boosting mechanism on a rare-pattern graph. The resulting approach leverages complementary notions of abnormality: structural deviations captured by graph reconstruction and symbolic rarity captured by mined patterns.
Experimental results on provenance traces from the DARPA TC program demonstrate that rare-pattern guidance consistently strengthens reconstruction-based scoring and improves anomaly ranking in highly imbalanced settings. Beyond performance, the proposed framework retains interpretability by allowing high-risk processes to be associated with the rare patterns they support, providing actionable cues for investigation.
In future work, we plan to (i) integrate pattern evidence directly into the training objective through pattern-aware regularization or heterogeneous graph modeling, (ii) extend the framework to streaming or online settings with incremental pattern updates, and (iii) incorporate richer temporal and causal constraints to better capture multi-stage attack behaviors across diverse host environments.

\section*{Conflicts of Interest}
The authors declare no conflict of interest.
\bibliographystyle{IEEEtran}
\bibliography{sample}

@article{DBLP:journals/fgcs/BenabderrahmaneHVCR24,
author = {Sidahmed Benabderrahmane and Ngoc Hoang and others},
  title        = {Hack me if you can: Aggregating autoencoders for countering persistent
                  access threats within highly imbalanced data},
  journal      = {Future Gener. Comput. Syst.},
  volume       = {160},
  pages        = {926--941},
  year         = {2024},
  doi          = {10.1016/J.FUTURE.2024.06.050},
  bibsource    = {dblp computer science bibliography, https://dblp.org}
}

@article{dehankar2026machine,
  title={Machine Learning in Cyber Security: A Literature Review},
  author={Dehankar, Pooja and others},
  journal={Implementing Enter. Cybersecurity With AI},
  pages={181--200},
  year={2026},
  publisher={IGI Global Scientific Publishing}
}

@article{al2025deep,
  title={Deep learning-based intrusion detection system for detecting IoT botnet attacks: a review},
  author={Al-Shurbaji, Tamara and others},
  journal={IEEE Access},
  volume={13},
  pages={11792--11822},
  year={2025},
  publisher={IEEE}
}

@inproceedings{benabderrahmane2025apt,
  title={APT-LLM: Embedding-Based Anomaly Detection of Cyber Advanced Persistent Threats Using Large Language Models},
  author={Benabderrahmane, Sidahmed and others},
  booktitle={2025 13th International Symposium on Digital Forensics and Security (ISDFS)},
  pages={1--6},
  year={2025},
  organization={IEEE}
}

@article{benabderrahmane2025ranking,
  title={Ranking-enhanced anomaly detection using Active Learning-assisted Attention Adversarial Dual AutoEncoder},
  author={Benabderrahmane, Sidahmed and Cheney, James and Rahwan, Talal},
  journal={Scientific Reports},
  volume={15},
  number={1},
  pages={41602},
  year={2025},
  publisher={Nature Publishing Group UK London}
}

@article{benabderrahmane2025one,
  title={From one attack domain to another: Contrastive transfer learning with siamese networks for APT detection},
  author={Benabderrahmane, Sidahmed and Rahwan, Talal},
  journal={Knowledge-Based Systems},
  pages={114877},
  year={2025},
  publisher={Elsevier}
}

@inproceedings{benabderrahmane2025flagus,
  title={FLAGUS: Feedback-Driven Learning with GAN-Augmented Data for Enhanced Anomaly Detection via Uncertainty Sampling},
  author={Benabderrahmane, Sidahmed and Rawhan, Talal},
  booktitle={International Conference on Neural Information Processing},
  pages={501--516},
  year={2025},
  organization={Springer}
}

@inproceedings{al2026dynamic,
  title={A Dynamic Hybrid IDS for APT Detection in ICS Using Deep Neural Networks},
  author={Al-Sinayyid, Rohith Battula D Ali},
  booktitle={AI Revolution: Research, Ethics and Society: International Conference, AIR-RES 2025, Las Vegas, NV, USA, April 14--16, 2025, Proceedings, Part II},
  pages={187},
  year={2026},
  organization={Springer Nature}
}

@article{wang2025dynamic,
  title={A dynamic provenance graph-based detector for advanced persistent threats},
  author={Wang, Lin and others},
  journal={Expert Systems with Applications},
  volume={265},
  pages={125877},
  year={2025},
  publisher={Elsevier}
}

@article{wu2025angus,
  title={Angus: efficient active learning strategies for provenance based intrusion detection},
  author={Wu, Lin and others},
  journal={Cybersecurity},
  volume={8},
  number={1},
  pages={6},
  year={2025},
  publisher={Springer}
}

@article{luo2025conformal,
  title={Conformal load prediction with transductive graph autoencoders},
  author={Luo, Rui and Colombo, Nicolo},
  journal={Machine Learning},
  volume={114},
  number={3},
  pages={1--22},
  year={2025},
  publisher={Springer}
}

@article{ma2023survey,
  title        = {A Comprehensive Survey on Graph Anomaly Detection with Deep Learning},
  author       = {Ma, Xiaoxiao and others},
  journal      = {IEEE Transactions on Knowledge and Data Engineering},
  volume       = {35},
  number       = {12},
  pages        = {12012--12038},
  year         = {2023},
  month        = dec,
  doi          = {10.1109/TKDE.2021.3118815}
}

@inproceedings{wang2023deep,
  title={Deep contrastive one-class time series anomaly detection},
  author={Wang, Rui and Liu, Chongwei and Mou, Xudong and Gao, Kai and Guo, Xiaohui and Liu, Pin and Wo, Tianyu and Liu, Xudong},
  booktitle={Proc. of the 2023 SIAM SDM conference},
  pages={694--702},
  year={2023},
  organization={SIAM}
}

@article{mir2023gcn,
  title        = {Graph Anomaly Detection with Graph Convolutional Networks},
  author       = {Mir, Aabid A. and Zuhairi, Megat F. and others},
  journal      = {International Journal of Advanced Computer Science and Applications (IJACSA)},
  volume       = {14},
  number       = {11},
  pages        = {601--613},
  year         = {2023}
}

@article{kipf2016vgae,
  title        = {Variational Graph Auto-Encoders},
  author       = {Kipf, Thomas N. and Welling, Max},
  journal      = {arXiv preprint arXiv:1611.07308},
  year         = {2016}
}

@article{alshamrani2019apt,
  title        = {A Survey on Advanced Persistent Threats: Techniques, Solutions, Challenges, and Research Opportunities},
  author       = {Alshamrani, Adel and Myneni, Sowmya and Chowdhary, Ankur and Huang, Dijiang},
  journal      = {IEEE Communications Surveys \& Tutorials},
  volume       = {21},
  number       = {2},
  pages        = {1851--1877},
  year         = {2019},
  month        = apr,
  doi          = {10.1109/COMST.2019.2891891},
  note         = {Article 8606252}
}

@article{berrada2020baseline,
  title        = {A Baseline for Unsupervised Advanced Persistent Threat Detection in System-Level Provenance},
  author       = {Berrada, Ghita and Cheney, James and Benabderrahmane, Sidahmed and Maxwell, William and Mookherjee, Himan and Theriault, Alec and Wright, Ryan},
  journal      = {Future Generation Computer Systems},
  volume       = {108},
  pages        = {401--413},
  year         = {2020},
  month        = jul,
  doi          = {10.1016/j.future.2020.02.015}
}

@inproceedings{han2020unicorn,
  title        = {{UNICORN}: Runtime Provenance-Based Detector for Advanced Persistent Threats},
  author       = {Han, Xueyuan and Pasquier, Thomas and Bates, Adam and Mickens, James and Seltzer, Margo},
  booktitle    = {Network and Distributed Systems Security (NDSS) Symposium},
  year         = {2020},
  month        = feb,
  address      = {San Diego, CA, USA},
  doi          = {10.14722/ndss.2020.24046}
}

@inproceedings{manzoor2016streamspot,
  title        = {Fast Memory-Efficient Anomaly Detection in Streaming Heterogeneous Graphs},
  author       = {Manzoor, Emaad A. and Momeni, Sadegh and Venkatakrishnan, V. N. and Akoglu, Leman},
  booktitle    = {Proc. of the 22nd ACM SIGKDD conference},
  year         = {2016},
  pages        = {1035--1044},
  doi          = {10.1145/2939672.2939783}
}

@inproceedings{hossain2017sleuth,
  title        = {{SLEUTH}: Real-Time Attack Scenario Reconstruction from {COTS} Audit Data},
  author       = {Hossain, Md. Nafis and others},
  booktitle    = {Proceedings of the 26th USENIX Security Symposium (USENIX Security 17)},
  year         = {2017},
  pages        = {487--504}
}

@inproceedings{agrawal1994apriori,
  title        = {Fast Algorithms for Mining Association Rules},
  author       = {Agrawal, Rakesh and Srikant, Ramakrishnan},
  booktitle    = {Proceedings of the 20th International Conference on Very Large Data Bases (VLDB)},
  year         = {1994},
  pages        = {487--499}
}

@inproceedings{benabderrahmane2021vrarm,
  title        = {A Rule Mining-based Advanced Persistent Threats Detection System},
  author       = {Benabderrahmane, Sidahmed and Cheney, James and others},
  booktitle    = {Proceedings of the Thirtieth International Joint Conference on Artificial Intelligence (IJCAI-21)},
    year         = {2021},
  pages        = {3589--3595}
}

\end{document}